\documentclass{article}
\usepackage{ijcai25}

\usepackage{times}
\usepackage{soul}
\usepackage{url}
\usepackage[hidelinks]{hyperref}
\usepackage[utf8]{inputenc}
\usepackage[small]{caption}
\usepackage{graphicx}
\usepackage{amsmath}
\usepackage{amsthm}
\usepackage{booktabs}
\usepackage{algorithm}
\usepackage{algorithmic}
\usepackage[switch]{lineno}

\usepackage{amssymb}
\usepackage{array}
\usepackage{caption}
\usepackage[acronym]{glossaries} 
\glsdisablehyper 
\usepackage{hyperref}
\usepackage{lipsum}
\usepackage{multirow}
\usepackage{subcaption}
\usepackage{tabularx}
\usepackage{xcolor}
\usepackage{xspace}

% ...

\newacronym{ai}{AI}{Artificial Intelligence}
\newacronym{clip}{CLIP}{Contrastive Language–Image Pre-training}
\newacronym{fm}{FM}{Foundation Model}
\newacronym{icl}{ICL}{In-Context Learning}
\newacronym{marl}{MARL}{Multi-Agent Reinforcement Learning}
\newacronym[longplural={Markov Decision Processes}]{mdp}{MDP}{Markov Decision Process}
\newacronym{mcts}{MCTS}{Monte Carlo Tree Search}
\newacronym{nlp}{NLP}{Natural Language Processing}
\newacronym{lm}{LM}{Language Model}
\newacronym{llm}{LLM}{Large Language Model}
\newacronym{rl}{RL}{Reinforcement Learning}
\newacronym{vip}{VIP}{Value-Implicit Pre-Training}
\newacronym{vlm}{VLM}{Vision Language Model}

\newcommand{\ai}[0]{\gls{ai}\xspace}

\newcommand{\fm}[0]{\gls{fm}\xspace}
\newcommand{\fms}[0]{\glspl{fm}\xspace}

\newcommand{\llm}[0]{\gls{llm}\xspace}
\newcommand{\llms}[0]{\glspl{llm}\xspace}

\newcommand{\mdp}[0]{\gls{mdp}\xspace}
\newcommand{\mdps}[0]{\glspl{mdp}\xspace}
\newcommand{\nlp}[0]{\gls{nlp}\xspace}
\newcommand{\rl}[0]{\gls{rl}\xspace}
\newcommand{\vlm}[0]{\gls{vlm}\xspace}
\newcommand{\vlms}[0]{\glspl{vlm}\xspace}

\newcommand{\chatglmturbo}{ChatGLM-Turbo}
\newcommand{\clipmodel}{CLIP}
\newcommand{\codelama}{CodeLlama}
\newcommand{\codex}{Codex}

\newcommand{\flantfive}{FLAN-T5}
\newcommand{\geminipro}{Gemini-Pro}
\newcommand{\gptthree}{GPT-3}
\newcommand{\gptthreefive}{GPT-3.5}
\newcommand{\gptthreefiveturbo}{GPT-3.5-Turbo}
\newcommand{\gptfour}{GPT-4}
\newcommand{\gptfourv}{GPT-4V}
\newcommand{\llama}{Llama}
\newcommand{\llamatwo}{Llama 2}
\newcommand{\longchat}{LongChat}
\newcommand{\meerkat}{Meerkat}
\newcommand{\palm}{PaLM}
\newcommand{\vicuna}{Vicuna}
\newcommand{\vicunaonefive}{Vicuna-1.5}

\newcommand{\ft}{\checkmark\xspace}
\newcommand{\noft}{\ensuremath{\times}\xspace}
\newcommand{\peft}{\checkmark*\xspace}

\newcommand{\single}{single\xspace}
\newcommand{\multi}{multi\xspace}

\newcommand{\online}{online\xspace}
\newcommand{\offline}{offline\xspace}
\newcommand{\both}{both\xspace}

\newcommand{\p}{\ensuremath{\pi}\xspace}
\newcommand{\pg}{\ensuremath{\pi_{g}}\xspace}
\newcommand{\pl}{\ensuremath{\pi_{l}}\xspace}

\renewcommand{\v}{\ensuremath{v}\xspace}
\newcommand{\vg}{\ensuremath{v_g}\xspace}
\newcommand{\vl}{\ensuremath{v_l}\xspace}

\newcommand{\petraj}{\ensuremath{\tau_{\pi}}\xspace}

\newcommand{\rft}{rft\xspace}

\title{The Evolving Landscape of LLM- and VLM-Integrated Reinforcement Learning}

\newcounter{savecntr}
\newcounter{restorecntr}

\author{
Sheila Schoepp$^1$
\and
Masoud Jafaripour$^1$\setcounter{savecntr}{\value{footnote}}\footnote{Equal contribution.} 
\and
Yingyue Cao$^1$\setcounter{restorecntr}{\value{footnote}}%
                \setcounter{footnote}{\value{savecntr}}\footnotemark%
                \setcounter{footnote}{\value{restorecntr}} 
                \and
Tianpei Yang$^2$ 
\and
Fatemeh Abdollahi$^1$ 
\and
Shadan Golestan$^{3}$ 
\and
Zahin Sufiyan$^1$ 
\and
Osmar R. Zaiane$^{1,3}$ 
\And
Matthew E. Taylor$^{1,3}$\\
\affiliations
$^1$University of Alberta\\ 
$^2$Nanjing University\\
$^3$Alberta Machine Intelligence Institute\\
}

\begin{document}

\maketitle


\begin{abstract}

Reinforcement learning (RL) has shown impressive results in sequential decision-making tasks. Meanwhile, Large Language Models (LLMs) and Vision-Language Models (VLMs) have emerged, exhibiting impressive capabilities in multimodal understanding and reasoning. These advances have led to a surge of research integrating LLMs and VLMs into RL. In this survey, we review representative works in which LLMs and VLMs are used to overcome key challenges in RL, such as lack of prior knowledge, long-horizon planning, and reward design. We present a taxonomy that categorizes these LLM/VLM-assisted RL approaches into three roles: agent, planner, and reward. We conclude by exploring open problems, including grounding, bias mitigation, improved representations, and action advice. By consolidating existing research and identifying future directions, this survey establishes a framework for integrating LLMs and VLMs into RL, advancing approaches that unify natural language and visual understanding with sequential decision-making.

\end{abstract}


\section{Introduction}


\rl is an influential branch of machine learning that enables autonomous agents to learn sequential decision-making strategies through an iterative process of trial-and-error interaction with their environment. When integrated with deep neural networks, deep \rl has made breakthroughs in challenging domains such as games and robotics~\cite{schulman2017proximal,vinyals2019grandmaster}. Despite these advances, \rl still faces key challenges: reliance on human-designed rewards, sample inefficiency, poor generalization, and limited interpretability, hindering real-world deployment. These limitations motivate the exploration of novel techniques to enhance the capabilities of \rl, particularly in areas where conventional approaches fall short.


\llms represent a groundbreaking advancement in \ai, exhibiting unprecedented capabilities in natural language understanding, generation, and reasoning. 
By training large architectures --- often spanning billions or even trillions of parameters --- on internet-scale datasets, \llms such as {GPT-3}~\cite{brown2020language}
have demonstrated emergent capabilities that smaller models could not achieve. Leveraging these strengths, \llms are now applied to tasks that extend beyond conventional \nlp, spanning domains from healthcare to robotics \cite{ichter2022can,thirunavukarasu2023large}. Similarly, \vlms, which integrate visual perception with natural language understanding, can interpret and reason about images through language. Leveraging large-scale, aligned image-text training, \vlms like CLIP \cite{radford2021learning} can perform a variety of tasks, including image-text retrieval and classification. Other \vlms, such as PaLM-E~\cite{driess2023palm}, are designed to respond to natural language prompts, broadening their versatility to tasks such as image captioning, scene understanding, and visual question answering. Together, these \fms, specifically \llms and \vlms, have reshaped \ai by capturing nuanced, human-centric semantics across modalities, enabling flexible, human-aligned problem-solving based on their vast training data.


Integrating \llms and \vlms into the \rl framework promises a transformative leap
in how agents act and learn. While \rl is proficient at learning from trial-and-error, it typically lacks the broad world knowledge and powerful reasoning capabilities that \llms and \vlms can provide. When integrated with \rl, these models enhance agents' capabilities by supplying semantic understanding (\llms) or robust perception (\vlms), thereby improving data efficiency, generalization, and interpretability. In some cases, RL’s ability to continually refine behaviour through interactions with the environment can complement these \fms by providing supplemental training or richer context and improving their outputs.

Research in the area of \llms and \vlms is driving a rapid evolution. As a result, integration of \fms into \rl is also progressing swiftly, further expanding the boundaries of what \rl can accomplish. Despite prior work on integrating LLMs into RL, the field's fast pace demands continual analysis of emerging methods and applications. Furthermore, with the emergence of \llm agents and powerful VLMs --- a perspective not addressed by earlier surveys --- this survey complements existing work by introducing these new dimensions, expanding our understanding of how best to integrate \fms with \rl.


For this survey, we selectively examine peer-reviewed studies that employ pretrained \llms and large \vlms developed on or after June 2020 --- coinciding with the release of GPT-3, a notable milestone in \nlp due to its unprecedented scale and capabilities --- as a core methodological component. These \fms must employ a transformer‐based architecture (whether encoder-only, decoder-only, or encoder-decoder) and address sequential decision‐making tasks framed as \mdps. We highlight works that use rewards to optimize \rl or \llm/\vlm policies for improved sequential decision-making. Although \rl can fine-tune language models, our focus is on using \fms to enhance \rl, not simply improving the models themselves. We include a representative selection of papers meeting these criteria, acknowledging that some relevant studies are omitted due to space constraints.


In summary, the main contributions of this survey include: \textbf{(1)} A unifying \textit{taxonomy} that categorizes \fm functionalities in \rl into three key roles: \llm/\vlm as Agent, \llm/\vlm as Planner, and \llm/\vlm as Reward. \textbf{(2)} A \textit{review} of key works within each category, highlighting how they address key \rl challenges such as policy learning, long-horizon planning, and reward specification. \textbf{(3)} \textit{Future directions} that identify limitations in existing approaches and outline promising paths for \fm-\rl research.


\section{Preliminaries}


\subsection{Reinforcement Learning}

A \textbf{Markov Decision Process} (\mdp) is defined by the tuple $\left( S, A, T, R, \gamma \right)$, where $S$ is the set of states, $A$ is the set of actions, $T: S \times A \rightarrow P(S)$ is the transition probability function, $R: S \times A \rightarrow \mathbb{R}$ is the reward function,
and $\gamma \in [0,1]$ is the discount factor~\cite{sutton2018reinforcement}.

\textbf{Reinforcement Learning} (\rl) is a paradigm in which an agent learns through interactions with an environment, typically modelled as an \mdp~\cite{sutton2018reinforcement}. These interactions produce a \textit{trajectory} of states, actions, and rewards as the agent explores its surroundings. A central concept in \rl is the \textit{policy}, $\pi$, which maps states to actions (or distributions over actions), formally expressed as $\pi : S \to P(A).$ In some settings, this is further extended to a \textit{language-conditioned policy}, $\pi_l: S \times \mathcal{L} \to P(A)$, where
$\mathcal{L}$ represents the space of natural language instructions (e.g., sub-goals), allowing the agent to incorporate linguistic guidance into its decisions. Under a policy $\pi$, the \textit{value function}, $v : S \to \mathbb{R}$ (or $q : S \times A \to \mathbb{R}$), estimates the expected cumulative reward from a given state (or state-action pair).
A language-conditioned (action-)value function  can likewise be conditioned on an instruction $l$.


\subsection{Language Models}


\textbf{Large Language Models} (\llms) learn statistical patterns in text from large corpora, enabling them to predict the likelihood of word (or token) sequences in context. They often rely on transformer architectures, which use self-attention to capture token dependencies \cite{vaswani2017attention}. Transformer-based \llms include encoder-only models that mask part of the input and learn to predict the missing portion (useful for text understanding), decoder-only models that generate text by predicting the next token in a sequence (often used for text generation), and encoder-decoder models that encode input into a latent representation and then decode it (common for translation tasks).


\textbf{Vision Language Models} (\vlms) are multimodal, processing both visual and textual data, often relying on transformers. They can be categorized into encoder-decoder models that convert images and/or text into latent embeddings before generating output (used for tasks like captioning), dual-encoder models that embed images and text separately into a shared latent space (used for similarity matching and retrieval), and single encoder models that encode images and text jointly (used for tasks like visual question answering).


\subsection{Taxonomy}

\begin{figure}[t]
    \centering
    \includegraphics[width=\linewidth]{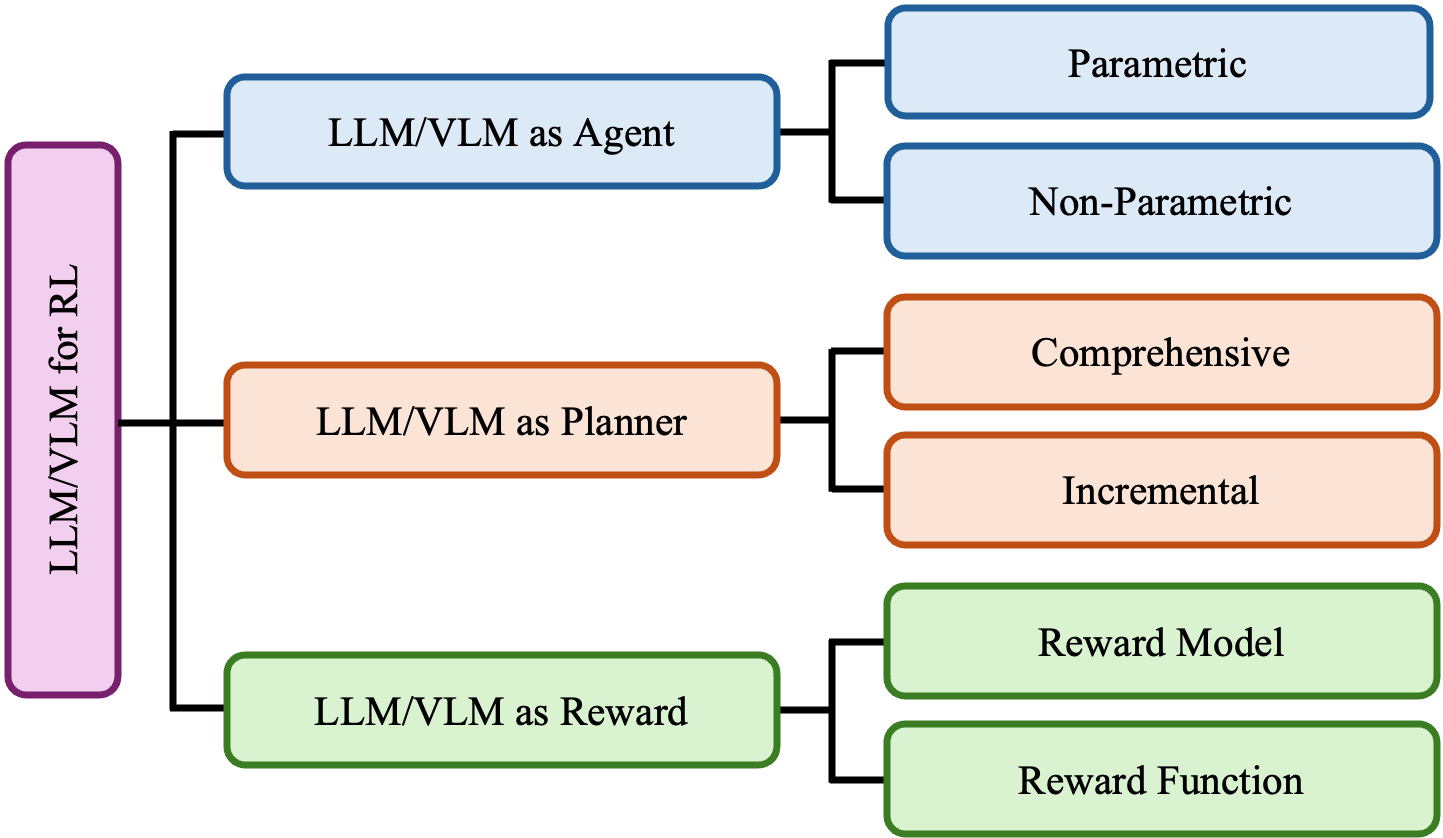}
    \caption{A taxonomy for \llm- and \vlm-assisted \rl.}
    \label{fig:taxonomy}
\end{figure}

Figure \ref{fig:taxonomy} presents a three‐part taxonomy to integrate \llms and \vlms into \rl, distinguishing three primary roles: (1) \llm/\vlm as Agent, where the \fm serves as a policy. These methods can either be parametric, fine‐tuning the \fm to generate task‐relevant outputs, or non-parametric, enriching the prompts with additional context. (2) \llm/\vlm as Planner, where the \fm generates sub‐goals for complex tasks. The \fm may produce a comprehensive sequence of sub‐goals in one pass or incrementally produce them (i.e., step by step), awaiting a signal of success or failure before generating the next sub‐goal. (3) \llm/\vlm as Reward, where the \fm shapes rewards by generating the reward function code to specify the reward or by serving as (or helping train) a reward model that outputs a scalar reward signal. Table 1 provides an overview of \fm-\rl methods, classified according to the taxonomy.

Some approaches do not fit into these three primary roles; instead, \llms/\vlms are integrated into \rl using alternative methods. For example, KALM~\cite{pang2024kalm} uses the \fm as a world model to generate ``imaginary" trajectories. Lai and Zang~\shortcite{lai2024sample} use an \fm to identify and emphasize higher-quality trajectories. MaestroMotif~\cite{klissarov2024maestromotif} and LAST~\cite{fu2024language} guide hierarchical \rl by discovering and coordinating skills.

In subsequent sections, we examine the three primary categories in our taxonomy, investigating the distinct ways that \llms and \vlms can be integrated into and benefit \rl.


\begin{table*}[th!]
\centering
\tiny
\label{tab:main_table}
    \begin{tabular}{%
        >{\centering\arraybackslash}m{0.1cm}
        >{\centering\arraybackslash}m{0.1cm}
        l                                  
        l                                  
        c     
        l     
        c    
        c  
        c  
        l  
        l   
        c
        }
        
        \toprule
        
        & 
        &
        & \multicolumn{3}{c}{\textbf{Foundation Model (FM)}}
        & \multicolumn{4}{c}{\textbf{Reinforcement Learning (RL)}}
        & 
        &
        \\
        \cmidrule(lr){4-6}\cmidrule(lr){7-10}
        
        &
        & \textbf{Citation}
        & \textbf{Model(s)}
        & \textbf{FT}
        & \textbf{Role}
        & \textbf{Agent}
        & \textbf{Task}
        & \textbf{Setting}
        & \textbf{Role}
        & \textbf{Metrics}
        & \textbf{Code}
        \\
        
        \midrule
        
        \multirow{14}{*}{\rotatebox[origin=c]{90}{LLM/VLM as Actor}}
        & \multirow{6}{*}{\rotatebox[origin=c]{90}{Parametric}} 
        & AGILE~\cite{he2024agile}
        & \meerkat, \vicunaonefive
        & \peft
        & act, ref
        & \single
        & -
        & \online
        & \pl, \vl, \rft
        & acc, rew
        & \href{https://github.com/bytarnish/AGILE}{Link}
        \\
        
        &
        & Retroformer~\cite{yao2024retroformer}
        & \gptthree, \gptfour, \longchat
        & \ft
        & act, cr, ref
        & \single
        & -
        & \offline
        & \pl, \rft
        & sr, se
        & -
        \\
        
        &
        & TWOSOME~\cite{tan2024true}
        & \llama
        & \peft
        & act
        & \single
        & -
        & \online
        & \pl, \vl, \rft
        & sr, rew, gen, se
        & \href{https://github.com/WeihaoTan/TWOSOME}{Link}
        \\

        &
        & POAD~\cite{wen2024reinforcing}
        & \codelama, \llamatwo
        & \peft
        & act
        & \single
        & -
        & \online
        & \pl, \vl, \rft
        & rew, gen, se
        & \href{https://github.com/morning9393/ADRL}{Link}
        \\

        &
        & GLAM~\cite{carta2023grounding}
        & \flantfive
        & \ft
        & act
        & \single
        & -
        & \online
        & \pl, \vl, \rft
        & se, gen
        & \href{https://github.com/flowersteam/Grounding_LLMs_with_online_RL}{Link}
        \\

        &
        & Zhai~\textit{et al.}~\cite{zhai2024fine}
        & LLaVA‑v1.6‑Mistral
        & \peft
        & act
        & \single
        & -
        & \online
        & \pl, \vl, \rft
        & sr
        & \href{https://github.com/RL4VLM/RL4VLM}{Link}
        \\

        \cmidrule{2-12}
        
        & \multirow{7}{*}{\rotatebox[origin=c]{90}{Non-Parametric}} 
        & ICPI~\cite{brooks2023large}
        & \codex
        & \noft
        & act, wm, ref
        & \single
        & -
        & \online
        & \pl, \vl, \petraj
        & rew, gen
        & \href{https://github.com/ethanabrooks/icpi}{Link}
        \\

        &
        & Reflexion~\cite{shinn2023reflexion}
        & \gptthree, \gptthreefiveturbo, \gptfour
        & \noft
        & act, eval, cr, ref
        & \single
        & -
        & \online
        & \pl, \petraj
        & sr, acc
        & \href{https://github.com/noahshinn/reflexion}{Link}
        \\
        
        &
        & REMEMBERER~\cite{zhang2023large}
        & \gptthreefive
        & \noft
        & act
        & \single
        & -
        & \online
        & \pl, \vl, \petraj
        & sr, rob
        & \href{https://github.com/OpenDFM/Rememberer}{Link}
        \\
        
        &
        & ExpeL~\cite{zhao2024expel}
        & \gptthreefiveturbo, \gptfour
        & \noft
        & act, cr, ref
        & \single
        & -
        & \online
        & \pl, \petraj
        & sr, gen
        & \href{https://github.com/LeapLabTHU/ExpeL}{Link}
        \\

        &
        & RLingua~\cite{chen2024rlingua}
        & \gptfour
        & \noft
        & act, plan, ref
        & \single
        & -
        & \online
        & \pg, \vg, \petraj
        & sr, se
        & -
        \\
        
        &
        & Xu \textit{et al.}~\cite {xu2024language}
        & \gptthreefiveturbo
        & \noft
        & act, o
        & \multi
        & -
        & \online
        & \pl, \vl
        & sr, rob
        & -
        \\

        &
        & LangGround~\cite{li2024language}
        & \gptfour
        & \noft
        & act, o
        & \multi
        & -
        & \online
        & \pl
        & sr, gen, se, int
        & \href{https://github.com/romanlee6/langground}{Link}
        \\

        \midrule
        
        \multirow{13}{*}{\rotatebox[origin=c]{90}{LLM/VLM as Planner}}
        & \multirow{7}{*}{\rotatebox[origin=c]{90}{Comprehensive}} 
        & SayTap~\cite{tang2023saytap}
        & \gptfour
        & \noft
        & plan
        & \single
        & \multi
        & \online
        & \pg, \vg
        & sr, acc
        & -
        \\

        &
        & LgTS~\cite{shukla2024lgts}
        & \llamatwo
        & \noft
        & plan
        & \single
        & \multi
        & \online
        & \pg, \vg
        & sr, se
        & -
        \\
        
        &
        & PSL~\cite{dalal2024plan}
        & \gptfour
        & \noft
        & plan, o
        & \single
        & \single
        & \online
        & \p, \v
        & sr, gen, se
        & \href{https://github.com/planseqlearn/planseqlearn}{Link}
        \\
        
        &
        & LLaRP~\cite{szot2024large}
        & \llama
        & \noft
        & plan, o 
        & \single
        & \multi
        & \online
        & \pg, \vg
        & sr, gen, rob, se
        & \href{https://github.com/apple/ml-llarp}{Link}
        \\
        
        &
        & LMA3~\cite{colas2023augmenting}
        & \gptthreefiveturbo
        & \noft
        & plan, rew, eval, o
        & \single
        & \multi
        & \online
        & \pg
        & gen, exp
        & -
        \\
        
        &
        & When2Ask~\cite{hu2024enabling}
        & \vicuna
        & \noft
        & plan
        & \single
        & \single
        & \online
        & \p, \v
        & sr
        & \href{https://github.com/ZJLAB-AMMI/LLM4RL}{Link}
        \\

        &
        & Inner Monologue~\cite{huang2022inner}
        & \gptthree, \palm
        & \noft
        & plan, cr, ref
        & \single
        & \multi
        & -
        & o
        & sr, rob, al
        & -
        \\

        \cmidrule{2-12}
        
        & \multirow{5}{*}{\rotatebox[origin=c]{90}{Incremental}} 
        & SayCan~\cite{ichter2022can}
        & \palm
        & \noft
        & plan
        & \single
        & \multi
        & \both
        & \pl, \vl, \rft
        & sr, rob
        & \href{https://github.com/google-research/google-research/tree/master/saycan}{Link}
        \\
        
        &
        & LLM4Teach~\cite{zhou2023large}
        & \chatglmturbo, \vicuna
        & \noft
        & plan
        & \single
        & \single
        & \online
        & \p, \v
        & sr, se
        & \href{https://github.com/ZJLAB-AMMI/LLM4Teach}{Link}
        \\
        
        &
        & AdaRefiner~\cite{zhang2024adarefiner}
        & \llamatwo, \gptfour
        & \peft
        & plan, cr, ref, o
        & \single
        & \multi
        & \online
        & \pl, \vl, \petraj
        & sr, rew, gen, exp
        & \href{https://github.com/PKU-RL/AdaRefiner}{Link}
        \\

        &
        & BOSS~\cite{zhang2023bootstrap}
        & \llama
        & \noft
        & plan, o
        & \single
        & \multi
        & \both
        & \pl, \vl, \rft, \petraj
        & sr, gen, rob, se
        & -
        \\
        
        &
        & Text2Motion~\cite{lin2023text2motion}
        & \codex, \gptthreefive
        & \noft
        & plan, o
        & \single
        & \multi
        & \offline
        & \p, \v
        & sr, gen, int
        & -
        \\

        \midrule
        
        \multirow{10}{*}{\rotatebox[origin=c]{90}{LLM/VLM as Reward}}
        & \multirow{3}{*}{\rotatebox[origin=c]{90}{Function}} 
        & Text2Reward~\cite{xie2024text2reward}
        & \gptfour
        & \noft
        & rew, ref
        & \single
        & \single
        & \online
        & \p
        & sr, se, al
        & \href{https://github.com/xlang-ai/text2reward}{Link}
        \\
        
        &
        & Zeng \textit{et al.}~\cite{zeng2024learning}
        & \gptfour
        & \noft
        & rew, eval, cr, ref
        & \single
        & \single
        & \online
        & \p, \petraj
        & sr, se
        & -
        \\
        
        &
        & Eureka~\cite{ma2024eureka}
        & \gptfour
        & \noft
        & rew, cr, ref
        & \single
        & \single
        & \online
        & \p, \v, \petraj
        & sr, gen, se, al
        & \href{https://github.com/eureka-research/Eureka}{Link}
        \\

        \cmidrule{2-12}

        & \multirow{6}{*}{\rotatebox[origin=c]{90}{Model}} 
        & Kwon \textit{et al.}~\cite{kwon2023reward}
        & \gptthree
        & \noft
        & rew
        & \single
        & \single
        & \online
        & \p, \v
        & acc, se, al
        & -
        \\
        
        &
        & PREDILECT~\cite{holk2024predilect}
        & \gptfour
        & \noft
        & rew, o
        & \single
        & \single
        & \online
        & \p
        & rew, se, al
        & -
        \\
        
        &
        & ELLM~\cite{du2023guiding}
        & \codex, \gptthree
        & \noft
        & rew, plan
        & \single
        & \multi
        & \online
        & \pl, \vl
        & sr, gen, se, exp
        & -
        \\
        
        &
        & RL-VLM-F~\cite{wang2024rl}
        & \geminipro, \gptfourv
        & \noft
        & rew, eval 
        & \single
        & \single
        & \online
        & \p, \v
        & sr, rew, se
        & \href{https://github.com/yufeiwang63/RL-VLM-F}{Link}
        \\

        &
        & VLM-RM~\cite{rocamonde2024vision}
        & \clipmodel
        & \noft
        & rew
        & \single
        & \single
        & \online
        & \p, \v
        & sr, al
        & \href{https://github.com/AlignmentResearch/vlmrm}{Link}
        \\
        
        &
        & MineCLIP~\cite{fan2022minedojo}
        & \clipmodel
        & \peft
        & rew, eval
        & \single
        & \multi
        & \online
        & \pl, \vl
        & sr, gen, se, al
        & \href{https://github.com/MineDojo/MineDojo}{Link}
        \\
        
        \bottomrule
      
    \end{tabular}
    \caption{A summary of approaches leveraging \fms, specifically \llms and \vlms, to enhance \rl, organized according to the taxonomy illustrated in Figure \ref{fig:taxonomy} and listed in order of mention. \textbf{FT (Fine-Tuning)} \checkmark (full fine-tuning), \checkmark* (parameter-efficient fine-tuning), and $\times$ (no fine-tuning). \textbf{FM Role} Generation of \textit{act} (actions), \textit{plan} (high-level plan), rew (reward function/model), \textit{wm} (world model), \textit{eval} (task success evaluations), \textit{cr} (critiques and improvement suggestions), \textit{ref} (refinement), and \textit{o} (other). \textbf{RL Agent} \textit{single} (single agent) and \textit{multi} (multi-agent). \textbf{RL Task} \textit{single} (single task) and \textit{multi} (multi-task). \textbf{RL Setting} \textit{online} (learning from real-time interactions), \textit{offline} (learning from precollected interactions), and \textit{both}. \textbf{RL Role} \p (policy learning), \pl (language-conditioned policy learning), \v (value function learning), \vl (language-conditioned value function learning), \petraj (policy execution to generate trajectories), \textit{ref} (reinforced fine-tuning), \textit{o} (other), and \textit{n/a} (no RL role). \textbf{Metrics} Improvements in \textit{acc} (accuracy $\uparrow$), \textit{sr} (success rate $\uparrow$), \textit{rew} (reward or return $\uparrow$), \textit{gen} (generalization $\uparrow$), \textit{rob} (robustness $\uparrow$), \textit{se} (sample efficiency $\uparrow$), \textit{exp} (exploration $\uparrow$), \textit{al} (alignment with humans $\uparrow$), and \textit{int} (interpretability $\uparrow$).}
\end{table*}


\section{LLM/VLM as Agent}

Language-based decision-making agents leverage the reasoning, planning, and generalization capabilities of \llms, enabling them to perform complex tasks in interactive environments. These agents interact with the environment, acting as decision-makers at each time step to generate context-based actions. Recent advances classify agents as parametric, fine-tuning \llms for dynamic adaptation, or non-parametric, using external resources and prompt engineering without altering the model. This section reviews key advances, focusing on fine-tuning, action decomposition, memory-driven strategies, and in-context learning for dynamic, multimodal environments.

\begin{figure}[tb!]
    \centering
    
    \begin{subfigure}[b]{0.49\linewidth}
        \centering
        \includegraphics[width=\linewidth]{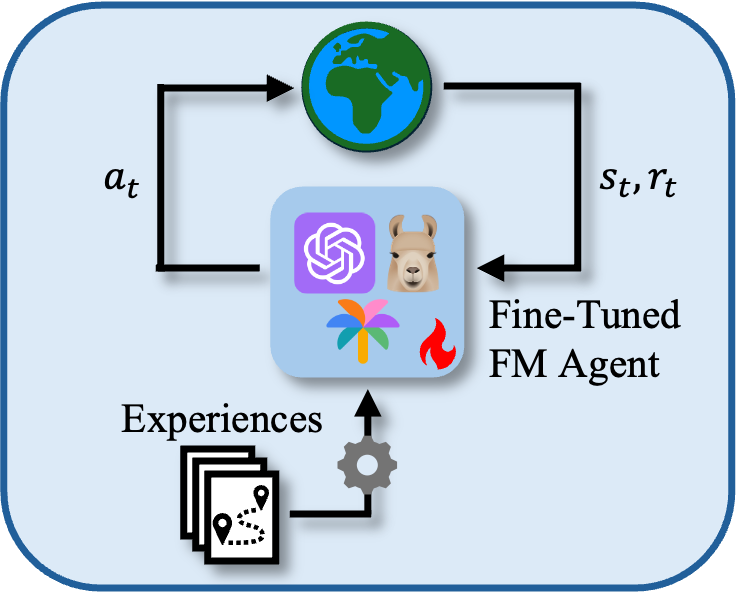}
        \caption{Parametric}
        \label{fig:lm_as_agent_parametric}
    \end{subfigure}
    \hfill
    \begin{subfigure}[b]{0.49\linewidth}
        \centering
        \includegraphics[width=\linewidth]{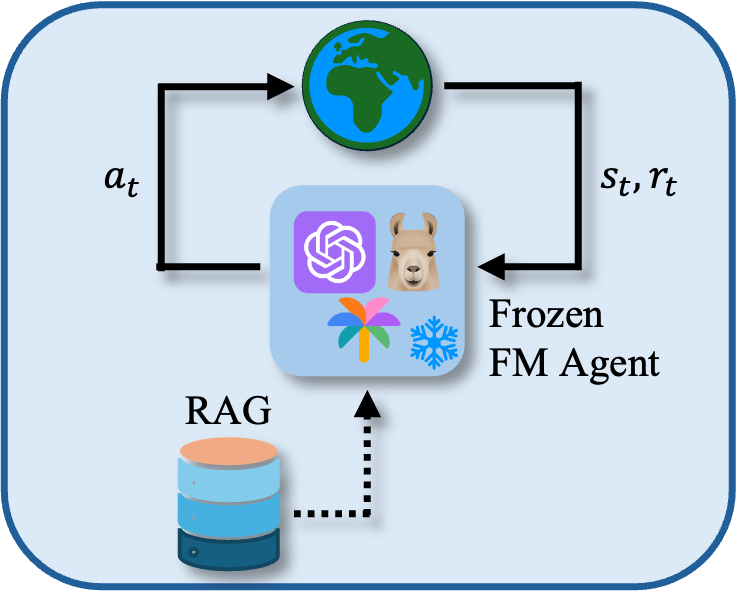}
        \caption{Non-parametric}
        \label{fig:lm_as_agent_non_parametric}
    \end{subfigure}
    
    \caption{LLM/VLM as Agent.}
    \label{fig:lm_as_agent}
\end{figure}


\subsection{Parametric}

Parametric \llm agents are decision-making models that fine-tune the internal parameters of \llms using experience datasets, as illustrated in Figure~\ref{fig:lm_as_agent_parametric}. This approach enables them to adapt their behaviour for specific tasks and environments, ensuring precise and context-aware decision-making. By leveraging \rl techniques such as policy optimization, action decomposition strategies, and value-based methods, these agents dynamically adjust their actions to align with specific task objectives.

For instance, AGILE~\cite{he2024agile} integrates memory, tools, and expert consultation within a modular framework, leveraging \rl to enhance reasoning and decision-making, achieving notable advancements over existing models in complex tasks. Outperforms the state-of-the-art \llm in specialized quality control benchmarks, demonstrating improved accuracy and adaptability. Similarly, Retroformer~\cite{yao2024retroformer} employs policy gradient optimization to iteratively refine prompts based on environmental feedback, achieving higher success rates in multi-step tasks. On the other hand, TWOSOME~\cite{tan2024true} improves sample efficiency and performance in interactive multi-step decision-making tasks by normalizing action probabilities and applying parameter-efficient fine-tuning to address alignment challenges between \llms and dynamic environments. Advanced methods further enhance parametric agents through innovative mechanisms. For example, POAD~\cite{wen2024reinforcing} decomposes actions into token-level decisions, addressing optimization complexity and enabling precise credit assignment in environments with large action spaces. GLAM~\cite{carta2023grounding} introduces functional grounding in textual environments, leveraging online \rl to align \llms with spatial and navigation tasks through step-by-step interaction and iterative learning. In vision-language tasks, fine-tuning frameworks combine chain-of-thought reasoning with \rl to enable agents to manage multimodal problems, demonstrating significantly enhanced visual-semantic understanding~\cite{zhai2024fine}.

Collectively, these approaches demonstrate that parametric \llm agents using \rl techniques, including policy optimization, action decomposition, and functional grounding, achieve superior adaptability, sample efficiency, and performance.


\subsection{Non-parametric}

Non-parametric \llm agents rely on the inherent reasoning and generalization capabilities of \llms, as shown in Figure~\ref{fig:lm_as_agent_non_parametric}, while keeping the \llm agent frozen and without altering its internal parameters. These agents leverage external resources and datasets, such as Retrieval Augmented Generation (RAG), enrich the task context, and use prompt engineering techniques during inference to guide decision-making, as recent works exemplify.

For example, ICPI~\cite{brooks2023large} implements policy iteration in \llms using in-context learning, where Q-values are computed via rollouts and iteratively refined. This approach, tested in six \rl domains, demonstrates the potential of \llms as both world models and policies, enabling scalable improvements without fine-tuning. Reflexion~\cite{shinn2023reflexion} introduces verbal reinforcement, where \llms generate and store self-reflective feedback in an episodic memory buffer to improve decision-making. This method enhances long-horizon decision-making, multi-step reasoning, and code generation, achieving state-of-the-art accuracy in function synthesis and logical inference. Similarly, REMEMBERER~\cite{zhang2023large} incorporates a persistent experience memory, allowing \llms to learn from past successes and failures in interactive environments without modifying parameters. By integrating \rl with experience memory, it improves adaptability and robustness in sequential reasoning and goal-oriented decision-making. Building on these ideas, ExpeL~\cite{zhao2024expel} introduces experiential learning, enabling \llms to autonomously collect, abstract, and apply knowledge from past tasks. This method enhances sequential decision-making and transfer learning, offering a resource-efficient alternative to fine-tuning.

Beyond general decision-making, non-parametric \llm agents have also been explored in domain-specific applications, including robotic manipulation and strategic multi-agent collaboration. RLingua~\cite{chen2024rlingua} improves sample efficiency in \rl for robotic manipulation by leveraging \llm-generated rule-based controllers as priors and integrating prior knowledge into policy learning through prompts. This approach enhances performance in sparse-reward tasks, achieving high success rates in both simulated and real-world environments with effective Sim2Real transfer. Werewolf~\cite{xu2024language} combines \llm-driven action candidate generation with \rl to mitigate intrinsic biases and enhance strategic decision-making. By integrating deductive reasoning and \rl, this framework enables agents to achieve human-level performance in unbounded communication and decision spaces. Similarly, LangGround~\cite{li2024language} aligns MARL agents’ communication with human language by grounding it in synthetic data from embodied \llms. This method facilitates zero-shot generalization in ad-hoc teamwork, improving communication emergence, interpretability, and task performance with unseen teammates. These studies illustrate that non-parametric \llm agents, by leveraging in-context learning, memory integration, self-reflection, and structured experience retrieval, can enhance reasoning, decision-making, and adaptability across diverse tasks, achieving state-of-the-art performance without requiring parameter updates.

\subsection{Discussion}
The integration of \llms/\vlms as decision-making agents highlights the strengths and limitations of parametric and non-parametric approaches. Parametric agents excel in task-specific adaptability and alignment via fine-tuning and \rl but face scalability and computational challenges in dynamic environments. Non-parametric agents leverage in-context learning and memory-driven reasoning for generalization and scalability without fine-tuning but struggle with long-term planning and complex modelling. These paradigms complement each other, with parametric methods providing precision and non-parametric approaches ensuring efficiency. Hybrid frameworks combining lightweight fine-tuning with advanced memory mechanisms can enhance \llm agents' robustness and adaptability in complex environments.


\section{LLM/VLM as Planner}

With extensive knowledge and strong reasoning capabilities, \fms can generate high-level plans that address \rl's struggles with complex multi-step tasks by decomposing them into sub-goals. Integrating \fms allows \rl agents focus on shorter-horizon control, improving sample efficiency when rewards are sparse or dependencies are intricate. Recent work suggests \fms provide powerful priors for \rl, though their planning ability remains heavily debated \cite{kambhampati2024llms}. We examine approaches that use \fms for plan generation in \rl, grouping them into two categories: comprehensive, where all sub-goals are planned upfront, and incremental, where sub-goals are generated step by step.

\begin{figure}[tb!]
    \centering
    
    \begin{subfigure}[b]{0.49\linewidth}
        \centering
        \includegraphics[width=\linewidth]{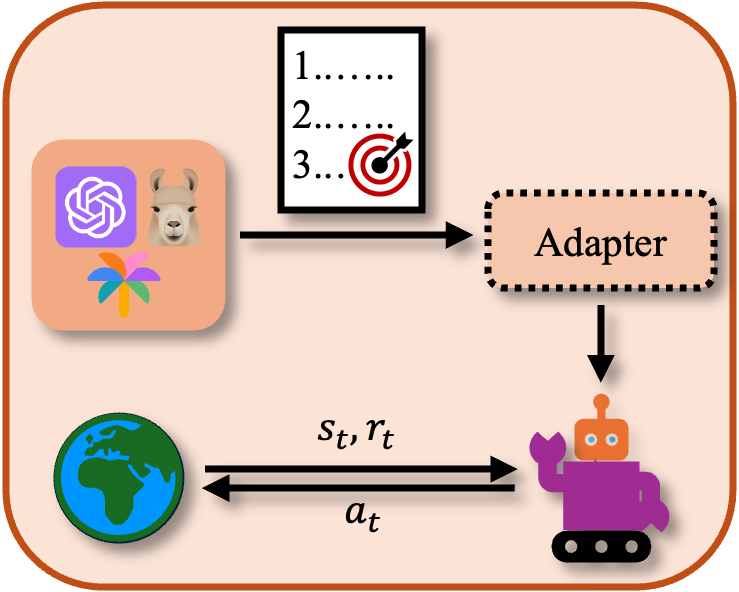}
        \caption{Comprehensive Planning}
        \label{fig:comprehensive_planning}
    \end{subfigure}
    \hfill
    \begin{subfigure}[b]{0.49\linewidth}
        \centering
        \includegraphics[width=\linewidth]{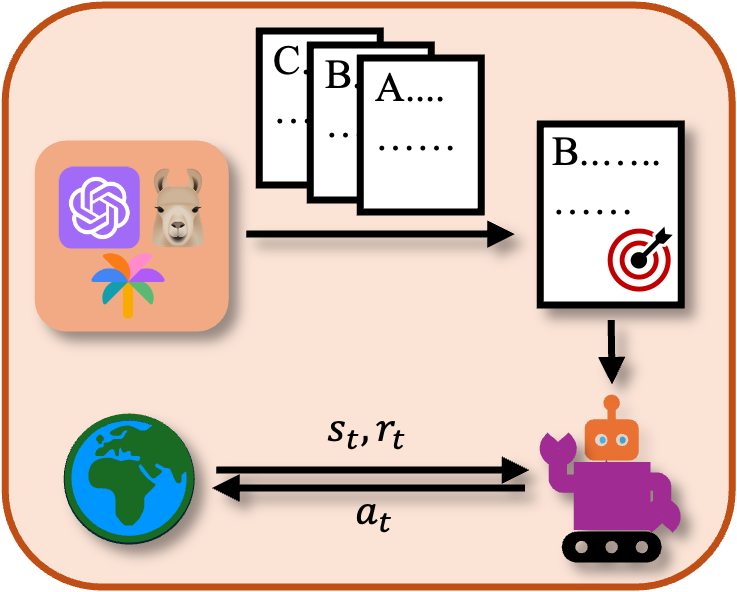}
        \caption{Incremental Planning}
        \label{fig:incremental_planning}
    \end{subfigure}
    
    \caption{LM as Planner.}
    \label{fig:lm_as_planner}
\end{figure}


\subsection{Comprehensive Planning}

\fms can generate a complete plan specifying sequential sub-goals for the agent to execute, as shown in Figure~\ref{fig:comprehensive_planning}. 
\fms inject their extensive knowledge into the planning process, they break down complex tasks into a sequence of achievable steps, freeing \rl agents from learning complex tasks from scratch and reducing overall training demands. When bridging the natural language plan to executable actions taken by the low-level controller, an advantage of using \fms is that the output of \fms can be structured based on actual needs.
For example, SayTap~\cite{tang2023saytap} uses foot contact patterns as a compact interface between language instructions and low-level quadruped control. An \llm outputs textual binary signals defining each leg's contact pattern, which an \rl policy is trained to follow. This demonstrates how high-level language commands can be translated into fine-grained control signals.
In addition to simpler binary-based control, some tasks may benefit from a skill library.
LMA3~\cite{colas2023augmenting} uses an \llm to evaluate and validate an agent’s performance on various goals, then treats the shortest action sequence from each successful execution as a skill.
LMA3 then leverages this growing skill library to chain short sequences into larger plans for solving a complex goal. However, its reliance on previously discovered action sequences limits its generalization.
Different from LMA3,  PSL~\cite{dalal2024plan} leverages the \llm to decompose the long-horizon natural language task into specially formatted language sub-goals. Each sub-goal contains lists of targeted regions for the robot to reach a termination stage, which is demanded by the motion planning module to plan to move the robot and a reinforcement learning policy learned to control. PSL removes the need for a pre-defined skill library and hence improves learning efficiency and generalization ability.

However, since the quality of the plan highly depends on the \fms, the initial plans may not be perfect, and execution failures might occur partway through. Appropriate adjustments and modifications to the plans generated by \fms can improve the correctness of the plan and hence improve the overall performance.
For example, Inner Monologue~\cite{huang2022inner} uses three types of feedback to update its plan in real-time. It collects binary feedback from a success detector after task accomplishment, visual to textual feedback from a scene detector during execution, and it is allowed to request a human or a Visual Question Answer model for feedback on questions he asked during execution. This dynamic re-planning skill 
improves completion rate and flexibility.
To avoid querying \llms after each failure execution and reduce the querying cost of \llms, LgTS~\cite{shukla2024lgts} uses \llms to generate multiple candidate sub-goal sequences before execution. It arranges them into a directed acyclic graph and employs an \rl Agent to explore the graph for the optimal path and learn the policy through a Teacher-Student learning strategy, speeding up learning and improving the sample efficiency.


\subsection{Incremental Planning}

Incremental Planning, as illustrated in Figure~\ref{fig:incremental_planning}, is another way for \fms to guide the agent, providing step-by-step guidance for actions. Querying \fms at every step incurs higher resource consumption costs; these approaches carefully determine when and how to query \fms at execution time.

For example, SayCan~\cite{ichter2022can} generates multiple candidate sub-goals at each step, then estimates each sub-goal’s likelihood of success. Combining sub-goals with these feasibility checks effectively grounds the \llm’s plans in real-world constraints, helping the agent to achieve the main goal.
Similarly, LLM4Teach~\cite{zhou2023large} provides the agent with a set of suggested actions to execute. Initially, the agent is trained to follow the guidance of an \llm closely, but as the agent learns over time, its dependence on the \llm's suggestions decreases, allowing the agent to make independent decisions.

Papers adopting incremental planning also improve the quality of these sub-goals through accumulating experience from past trajectories. 
For example, AdaRefiner~\cite{zhang2024adarefiner} enhances the agent's execution and understanding of \llm guidance by introducing a secondary \llm to evaluate the alignment of the agent's execution process and the guidance of \llm. The feedback from the agent, combined with evaluation scores from the secondary \llm, is then used to fine-tune the primary \llm, enabling it to provide better guidance in subsequent iterations. 
Similarly, BOSS~\cite{zhang2023bootstrap} learns from past trajectories but eliminates the need for a critic \llm. Instead, the guidance \llm continuously accumulates new skills demonstrated by the agent and adds them to a skill library. 
While summarizing and analyzing experiences from past trajectories could improve planning ability, simulating future trajectories can also contribute to better decision-making. 

Instead of using only natural language input, LLaRP~\cite{szot2024large} integrates a frozen \llm with a pre-trained vision encoder to process textual instructions and egocentric visual frames. LLaRP trains vision encoder and action decoder using online \rl, improving the robustness and generalization over the new environment.
A unique example is Text2Motion~\cite{lin2023text2motion}, which combines both Comprehensive and Incremental Planning, ensuring efficiency and correctness. Initially, Text2Motion employs an \llm to generate a comprehensive plan, encompassing all the steps for the agent to execute. If a planning failure arises during execution, Text2Motion employs the \llm to generate the actions incrementally. 

\subsection{Discussion}

LLM/VLM-based planning uses the common knowledge in \fms to break down complex tasks into simpler subtasks, improving learning efficiency. It is particularly effective in human-centric environments, where plans in natural language benefit from common-sense reasoning. Comprehensive planning can be more efficient but is riskier in dynamic settings, while incremental planning enables real-time feedback and adaptation but increases computational overhead. Balancing these approaches and translating model-generated plans into actionable steps that generalize across environments remain key challenges.


\section{LLM/VLM as Reward}

Designing effective reward signals remains a central challenge in \rl, requiring domain knowledge and trial-and-error tuning. While methods like preference-based learning, inverse \rl, and labelled datasets help, they still rely heavily on human input. Recent advances leverage \llms and \vlms for automating reward design by having them interpret textual descriptions and process visual inputs. These \llm/\vlm as Reward approaches generally fall into two categories: generating explicit reward functions, or serving as (or aiding the learning of) a reward model.

\begin{figure}[tb!]
    \centering
    
    \begin{subfigure}[b]{0.49\linewidth}
        \centering
        \includegraphics[width=\linewidth]{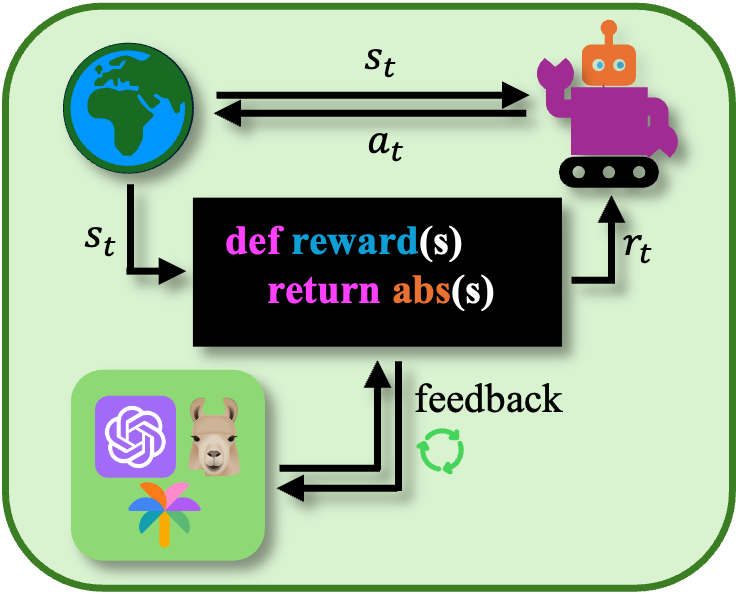}
        \caption{Reward Function}
        \label{fig:reward_function}
    \end{subfigure}
    \hfill
    \begin{subfigure}[b]{0.49\linewidth}
        \centering
        \includegraphics[width=\linewidth]{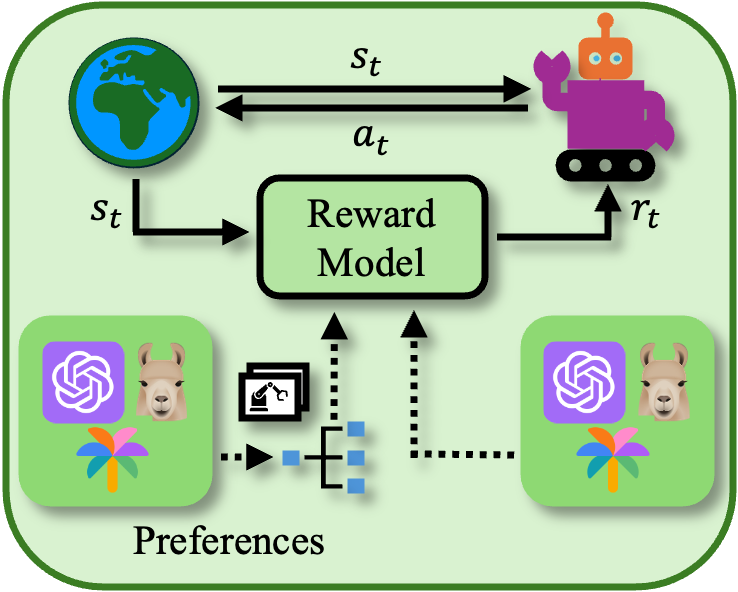}
        \caption{Reward Model}
        \label{fig:reward_model}
    \end{subfigure}
    
    \caption{LLM/VLM as Reward}
    \label{fig:lm_as_reward}
\end{figure}


\subsection{Reward Function}

Leveraging \llms to design reward functions addresses a significant bottleneck in \rl. It reduces human reward engineering effort, facilitates the discovery of novel reward components, and yields interpretable code. Providing a Pythonic environment abstraction as initial context and prompting an \llm iteratively generate and improve reward functions using natural language (as illustrated in Figure \ref{fig:reward_function}). These benefits are especially valuable for high‐dimensional or otherwise complex tasks.

Reward function approaches primarily differ in how they trigger refinements and the type of natural language feedback they incorporate.
For example, in Text2Reward~\cite{xie2024text2reward}, an \llm refines the reward function code until it executes successfully. After training an \rl policy, non-expert users can observe the learned policy and provide linguistic feedback on suboptimal behaviours, prompting further \llm refinements to the reward function.
Zeng \textit{et al.}~\shortcite{zeng2024learning} use an \llm to identify key behavioural features (to promote or discourage) and propose an initial reward function parameterization. The \llm iteratively refines this parameterization by ranking trajectories from executions of the trained policy, shaping the reward function toward desirable behaviours.
Meanwhile, Eureka~\cite{ma2024eureka} uses an evolutionary search strategy. At each iteration, an \llm generates multiple candidate reward functions, trains a policy for each, and then selects the best-performing policy for further refinement. This selection is guided by both policy performance and reward-function component metrics.
All three approaches produce reward functions that match or surpass those designed by human experts, and are readily extended to novel tasks with minimal human intervention.


\subsection{Reward Model}

As illustrated in Figure~\ref{fig:reward_model}, \glsentryfullpl{fm} can specify reward models in two key ways. First, \llms can serve as proxy reward models by mapping textual descriptions of desired behaviours directly to scalar rewards. Second, a separate reward model can be learned by leveraging \llms or \vlms to incorporate preference feedback on agent trajectories or by combining textual instructions with visual observations in \vlms to produce more robust and visually grounded reward models.


Kwon \textit{et al.}~\shortcite{kwon2023reward} use an \llm as a proxy reward model, using natural language descriptions of desired behaviours and textual trajectory summaries to generate a binary reward signal that guides policy learning. Strikingly, their straightforward approach performs nearly as effectively as ground truth while removing the need for large, curated datasets of preference labels or expert demonstrations.
PREDILECT~\cite{holk2024predilect} builds on preference-based \rl, allowing human raters to specify both their preferred trajectory and the reasons for their choice. Using these explanations, an \llm extracts key trajectory subsequences and incorporates them into the reward-learning objective via regularization, giving more weight to segments marked as ``good" or ``bad". This targeted influence mitigates causal confusion by directing the model's attention to the true causal factors underlying human preferences. 
ELLM~\cite{du2023guiding} improves exploration in \rl by prompting an \llm with a textual ``caption" of the agent’s state to generate sub-goals. The agent is rewarded for achieving these sub-goals via a semantic-similarity measure between its transition caption (action and resulting state) and the suggested sub-goal, with a novelty bias that rewards each sub-goal only once per episode. ELLM shifts naive novelty-driven exploration toward semantically guided skill discovery, yielding more human-like behaviours and faster task learning.


Text-based reward design often fails for visually complex tasks, where nuanced details cannot be fully expressed in words. RL-VLM-F~\cite{wang2024rl} overcomes this limitation by leveraging a large \vlm without requiring any human annotation, using it to rank pairs of images (observations) based on their alignment with a natural language task description. These pairwise preferences train a visually-grounded reward model, enabling robust reward design for tasks with intricate visual observations.
VLM-RM~\cite{rocamonde2024vision} and MineCLIP~\cite{fan2022minedojo} both leverage a large \vlm (CLIP) to scale \rl to tasks that are not easily specified using engineered reward functions but are easily described in natural language. VLM-RM targets continuous control problems by computing a direct scalar reward based on the cosine similarity between a textual goal embedding --- adjusted by subtracting a ``baseline prompt" embedding to reduce interference from irrelevant features --- and the agent's visual observation embedding. Notably, VLM-RM performance improves when environments are enhanced with more realistic visuals, better aligning with CLIP’s training distribution. MineCLIP similarly builds on CLIP but targets Minecraft’s open-ended environment, fine-tuning on 16-frame YouTube video segments paired with time-aligned text, yielding a dense reward signal that correlates the agent's recent frames with a free-form textual goal.

\subsection{Discussion}

\llm/\vlm as Reward approaches automate the generation of reward functions by translating textual descriptions into rewards for \rl agents. Their strong performance --- often matching or surpassing human-engineered and ground-truth rewards --- indicates that natural language effectively encodes and guides reward design for complex tasks. These approaches often face several constraints. They can be overly sensitive to prompt design, prone to hallucinations, or omit critical details. They also rely on simplified abstractions that fail to capture real-world complexity, raising concerns about scalability and reliability in more realistic settings.


\section{Future Directions}

Building on current methods and approaches, significant opportunities remain to advance this domain even further.


\subsection{Grounding}

\llms demonstrate strong capabilities in generating high-level plans, but they lack real-world experience, so their plans may not be executable for embodied agents such as robots~\cite{ichter2022can,dalal2024plan}. Current works solving the grounding problem by applying a bridging layer or verification module between the high-level plan and the low-level controller~\cite{dalal2024plan,huang2022inner} or by leveraging the value-function to ground the action~\cite{ichter2022can}. However, these methods share similar disadvantages: the external knowledge they rely on might introduce biases that negatively affect certain tasks. Another approach is to carefully design the plan's structure generated by \llms to fit the real-world requirement~\cite{tang2023saytap}, which also faces the problem of lacking generalization in diverse tasks and environments. Developing a more generalized and bias-free grounding method remains an important area for future research.


\subsection{Inherent Bias}

\llms and \vlms exhibit intrinsic biases rooted in their data sources, training procedures, and architectures, leading to suboptimal decisions. For example, an \llm can identify the Rock-Paper-Scissors Nash equilibrium --- playing each action equally --- yet still favour Rock, making it exploitable~\cite{xu2024language}. 
Few works target de-biasing, using techniques such as self-consistency and population-based training~\cite{xu2024language}, but only partially address the issue.
Meanwhile, implicit refinements or corrections of an LLM's outputs, through action values or environment feedback, have shown promise but remain largely confined to high-level task planning~\cite{huang2022inner,ichter2022can}. These limitations highlight the need for more robust and generalizable bias mitigation techniques, such as \rl-driven exploration, that can systematically expose and overcome these biases.


\subsection{Representation}

Integration of \llms into \rl is hindered by the need to convert rich numeric signals, such as raw sensor data and actions, into sequences of textual tokens, losing the nuanced semantic information required for precise control~\cite{du2023guiding,hu2023language}. KALM~\cite{pang2024kalm} addresses this limitation by replacing the \llm embedding and output layers with multilayer perceptron modules, enabling bidirectional translation between language goals and numeric trajectories. Building on KALM, a promising direction is to explore novel methods for modifying \llm architectures to fuse raw sensor data with language for joint multimodal representations or using \vlms to preserve rich feature representations while retaining language-based reasoning. A potentially powerful approach may combine \llms and \vlms, creating multimodal models capable of advanced language understanding, reasoning, decision-making, and visual perception --- paving the way for \rl agents to address complex tasks that demand richer representation.


\subsection{Action Advice}

Human-in-the-loop \rl, in which a human or human-simulating oracle provides real-time, action-level guidance (e.g., ``turn right," ``move forward") to the \rl agent, can significantly boost learning speed and performance in domains such as robotics, navigation, and games~\cite{rosenfeld2018leveraging,torrey2013teaching}. Recent advancements with \llms and \vlms show promise for providing similar guidance without direct human oversight. Instead of requiring a human to monitor the agent, these human-aligned models can serve as ``virtual oracles," issuing low-level instructions and removing the need for a human teacher. It is not necessary for these models to offer perfect advice; even occasional correctness can reduce the agent's exploration time~\cite{icarte2018advice}.


\section{Conclusion}

Research integrating \fms, particularly \llms and \vlms, with \rl is rapidly expanding. This survey introduces a taxonomy categorizing \fm-based methods into Agent, Planner, and Reward roles. We review studies in each role, highlighting how \fms can serve as parametric or non-parametric policies, generate comprehensive or incremental plans, or define rewards through a reward function or model. We discuss current limitations and propose future directions, aiming to clarify advancements and challenges in leveraging \fms for \rl and inspire further innovation.


\bibliographystyle{named}
\bibliography{ijcai25}

\begin{thebibliography}{}

\bibitem[\protect\citeauthoryear{Brooks \bgroup \em et al.\egroup }{2023}]{brooks2023large}
Ethan~A. Brooks, Logan Walls, et~al.
\newblock Large language models can implement policy iteration.
\newblock In {\em NeurIPS}, 2023.

\bibitem[\protect\citeauthoryear{Brown \bgroup \em et al.\egroup }{2020}]{brown2020language}
Tom~B. Brown, Benjamin Mann, et~al.
\newblock Language models are few-shot learners.
\newblock In {\em NeurIPS}, 2020.

\bibitem[\protect\citeauthoryear{Carta \bgroup \em et al.\egroup }{2023}]{carta2023grounding}
Thomas Carta, Cl{\'{e}}ment Romac, et~al.
\newblock Grounding large language models in interactive environments with online reinforcement learning.
\newblock In {\em ICML}, 2023.

\bibitem[\protect\citeauthoryear{Chen \bgroup \em et al.\egroup }{2024}]{chen2024rlingua}
Liangliang Chen, Yutian Lei, et~al.
\newblock Rlingua: Improving reinforcement learning sample efficiency in robotic manipulations with large language models.
\newblock {\em IEEE Robotics and Automation Letters}, 9(7), 2024.

\bibitem[\protect\citeauthoryear{Colas \bgroup \em et al.\egroup }{2023}]{colas2023augmenting}
C{\'{e}}dric Colas, Laetitia Teodorescu, et~al.
\newblock Augmenting autotelic agents with large language models.
\newblock In {\em CoLLAs}, 2023.

\bibitem[\protect\citeauthoryear{Dalal \bgroup \em et al.\egroup }{2024}]{dalal2024plan}
Murtaza Dalal, Tarun Chiruvolu, et~al.
\newblock Plan-seq-learn: Language model guided {RL} for solving long horizon robotics tasks.
\newblock In {\em ICLR}, 2024.

\bibitem[\protect\citeauthoryear{Driess \bgroup \em et al.\egroup }{2023}]{driess2023palm}
Danny Driess, Fei Xia, et~al.
\newblock Palm-e: An embodied multimodal language model.
\newblock In {\em ICML}, 2023.

\bibitem[\protect\citeauthoryear{Du \bgroup \em et al.\egroup }{2023}]{du2023guiding}
Yuqing Du, Olivia Watkins, et~al.
\newblock Guiding pretraining in reinforcement learning with large language models.
\newblock In {\em ICML}, 2023.

\bibitem[\protect\citeauthoryear{Fan \bgroup \em et al.\egroup }{2022}]{fan2022minedojo}
Linxi Fan, Guanzhi Wang, et~al.
\newblock Minedojo: Building open-ended embodied agents with internet-scale knowledge.
\newblock In {\em NeurIPS}, 2022.

\bibitem[\protect\citeauthoryear{Feng \bgroup \em et al.\egroup }{2024}]{he2024agile}
Peiyuan Feng, Yichen He, et~al.
\newblock Agile: A novel reinforcement learning framework of llm agents.
\newblock In {\em NeurIPS}, 2024.

\bibitem[\protect\citeauthoryear{Fu \bgroup \em et al.\egroup }{2024}]{fu2024language}
Haotian Fu, Pratyusha Sharma, et~al.
\newblock Language-guided skill learning with temporal variational inference.
\newblock In {\em ICML}, 2024.

\bibitem[\protect\citeauthoryear{Holk \bgroup \em et al.\egroup }{2024}]{holk2024predilect}
Simon Holk, Daniel Marta, et~al.
\newblock {PREDILECT:} preferences delineated with zero-shot language-based reasoning in reinforcement learning.
\newblock In {\em HRI}, 2024.

\bibitem[\protect\citeauthoryear{Hu and Sadigh}{2023}]{hu2023language}
Hengyuan Hu and Dorsa Sadigh.
\newblock Language instructed reinforcement learning for human-ai coordination.
\newblock In {\em ICML}, 2023.

\bibitem[\protect\citeauthoryear{Hu \bgroup \em et al.\egroup }{2024}]{hu2024enabling}
Bin Hu, Chenyang Zhao, et~al.
\newblock Enabling intelligent interactions between an agent and an {LLM:} {A} reinforcement learning approach.
\newblock {\em {RLJ}}, 3, 2024.

\bibitem[\protect\citeauthoryear{Huang \bgroup \em et al.\egroup }{2022}]{huang2022inner}
Wenlong Huang, Fei Xia, et~al.
\newblock Inner monologue: Embodied reasoning through planning with language models.
\newblock In {\em CoRL}, 2022.

\bibitem[\protect\citeauthoryear{Icarte \bgroup \em et al.\egroup }{2018}]{icarte2018advice}
Rodrigo~Toro Icarte, Toryn~Q. Klassen, et~al.
\newblock Advice-based exploration in model-based reinforcement learning.
\newblock In {\em Canadian {AI}}, 2018.

\bibitem[\protect\citeauthoryear{Ichter \bgroup \em et al.\egroup }{2022}]{ichter2022can}
Brian Ichter, Anthony Brohan, et~al.
\newblock Do as {I} can, not as {I} say: Grounding language in robotic affordances.
\newblock In {\em CoRL}, 2022.

\bibitem[\protect\citeauthoryear{Kambhampati \bgroup \em et al.\egroup }{2024}]{kambhampati2024llms}
Subbarao Kambhampati, Karthik Valmeekam, et~al.
\newblock Position: Llms can't plan, but can help planning in llm-modulo frameworks.
\newblock In {\em ICML}, 2024.

\bibitem[\protect\citeauthoryear{Klissarov \bgroup \em et al.\egroup }{2025}]{klissarov2024maestromotif}
Martin Klissarov, Mikael Henaff, et~al.
\newblock Maestromotif: Skill design from artificial intelligence feedback.
\newblock In {\em ICLR}, 2025.

\bibitem[\protect\citeauthoryear{Kwon \bgroup \em et al.\egroup }{2023}]{kwon2023reward}
Minae Kwon, Sang~Michael Xie, et~al.
\newblock Reward design with language models.
\newblock In {\em ICLR}, 2023.

\bibitem[\protect\citeauthoryear{Lai and Zang}{2024}]{lai2024sample}
Jinbang Lai and Zhaoxiang Zang.
\newblock Sample trajectory selection method based on large language model in reinforcement learning.
\newblock {\em {IEEE} Access}, 12, 2024.

\bibitem[\protect\citeauthoryear{Li \bgroup \em et al.\egroup }{2024}]{li2024language}
Huao Li, Hossein~Nourkhiz Mahjoub, et~al.
\newblock Language grounded multi-agent reinforcement learning with human-interpretable communication.
\newblock In {\em NeurIPS}, 2024.

\bibitem[\protect\citeauthoryear{Lin \bgroup \em et al.\egroup }{2023}]{lin2023text2motion}
Kevin Lin, Christopher Agia, et~al.
\newblock Text2motion: from natural language instructions to feasible plans.
\newblock {\em Autonomous Robots}, 47(8), 2023.

\bibitem[\protect\citeauthoryear{Ma \bgroup \em et al.\egroup }{2024}]{ma2024eureka}
Yecheng~Jason Ma, William Liang, et~al.
\newblock Eureka: Human-level reward design via coding large language models.
\newblock In {\em ICLR}, 2024.

\bibitem[\protect\citeauthoryear{Pang \bgroup \em et al.\egroup }{2024}]{pang2024kalm}
Jing-Cheng Pang, Si-Hang Yang, et~al.
\newblock Kalm: Knowledgeable agents by offline reinforcement learning from large language model rollouts.
\newblock In {\em NeurIPS}, 2024.

\bibitem[\protect\citeauthoryear{Radford \bgroup \em et al.\egroup }{2021}]{radford2021learning}
Alec Radford, Jong~Wook Kim, et~al.
\newblock Learning transferable visual models from natural language supervision.
\newblock In {\em ICML}, 2021.

\bibitem[\protect\citeauthoryear{Rocamonde \bgroup \em et al.\egroup }{2024}]{rocamonde2024vision}
Juan Rocamonde, Victoriano Montesinos, et~al.
\newblock Vision-language models are zero-shot reward models for reinforcement learning.
\newblock In {\em ICLR}, 2024.

\bibitem[\protect\citeauthoryear{Rosenfeld \bgroup \em et al.\egroup }{2017}]{rosenfeld2018leveraging}
Ariel Rosenfeld, Matthew~E. Taylor, et~al.
\newblock Leveraging human knowledge in tabular reinforcement learning: {A} study of human subjects.
\newblock In {\em IJCAI}, 2017.

\bibitem[\protect\citeauthoryear{Schulman \bgroup \em et al.\egroup }{2017}]{schulman2017proximal}
John Schulman, Filip Wolski, et~al.
\newblock Proximal policy optimization algorithms.
\newblock {\em arXiv preprint arXiv:1707.06347}, 2017.

\bibitem[\protect\citeauthoryear{Shinn \bgroup \em et al.\egroup }{2023}]{shinn2023reflexion}
Noah Shinn, Federico Cassano, et~al.
\newblock Reflexion: language agents with verbal reinforcement learning.
\newblock In {\em NeurIPS}, 2023.

\bibitem[\protect\citeauthoryear{Shukla \bgroup \em et al.\egroup }{2024}]{shukla2024lgts}
Yash Shukla, Wenchang Gao, et~al.
\newblock Lgts: Dynamic task sampling using llm-generated sub-goals for reinforcement learning agents.
\newblock In {\em AAMAS}, 2024.

\bibitem[\protect\citeauthoryear{Sutton and Barto}{2018}]{sutton2018reinforcement}
Richard~S. Sutton and Andrew~G. Barto.
\newblock {\em Reinforcement learning: An introduction}.
\newblock {MIT} Press, 2018.

\bibitem[\protect\citeauthoryear{Szot \bgroup \em et al.\egroup }{2024}]{szot2024large}
Andrew Szot, Max Schwarzer, et~al.
\newblock Large language models as generalizable policies for embodied tasks.
\newblock In {\em ICLR}, 2024.

\bibitem[\protect\citeauthoryear{Tan \bgroup \em et al.\egroup }{2024}]{tan2024true}
Weihao Tan, Wentao Zhang, et~al.
\newblock True knowledge comes from practice: Aligning large language models with embodied environments via reinforcement learning.
\newblock In {\em ICLR}, 2024.

\bibitem[\protect\citeauthoryear{Tang \bgroup \em et al.\egroup }{2023}]{tang2023saytap}
Yujin Tang, Wenhao Yu, et~al.
\newblock Saytap: Language to quadrupedal locomotion.
\newblock In {\em CoRL}, 2023.

\bibitem[\protect\citeauthoryear{Thirunavukarasu \bgroup \em et al.\egroup }{2023}]{thirunavukarasu2023large}
Arun~James Thirunavukarasu, Darren Shu~Jeng Ting, et~al.
\newblock Large language models in medicine.
\newblock {\em Nature medicine}, 29(8), 2023.

\bibitem[\protect\citeauthoryear{Torrey and Taylor}{2013}]{torrey2013teaching}
Lisa Torrey and Matthew~E. Taylor.
\newblock Teaching on a budget: agents advising agents in reinforcement learning.
\newblock In {\em AAMAS}, 2013.

\bibitem[\protect\citeauthoryear{Vaswani \bgroup \em et al.\egroup }{2017}]{vaswani2017attention}
Ashish Vaswani, Noam Shazeer, et~al.
\newblock Attention is all you need.
\newblock In {\em NeurIPS}, 2017.

\bibitem[\protect\citeauthoryear{Vinyals \bgroup \em et al.\egroup }{2019}]{vinyals2019grandmaster}
Oriol Vinyals, Igor Babuschkin, et~al.
\newblock Grandmaster level in starcraft {II} using multi-agent reinforcement learning.
\newblock {\em Nature}, 575(7782), 2019.

\bibitem[\protect\citeauthoryear{Wang \bgroup \em et al.\egroup }{2024}]{wang2024rl}
Yufei Wang, Zhanyi Sun, et~al.
\newblock {RL-VLM-F:} reinforcement learning from vision language foundation model feedback.
\newblock In {\em ICML}, 2024.

\bibitem[\protect\citeauthoryear{Wen \bgroup \em et al.\egroup }{2024}]{wen2024reinforcing}
Muning Wen, Ziyu Wan, et~al.
\newblock Reinforcing llm agents via policy optimization with action decomposition.
\newblock In {\em NeurIPS}, 2024.

\bibitem[\protect\citeauthoryear{Xie \bgroup \em et al.\egroup }{2024}]{xie2024text2reward}
Tianbao Xie, Siheng Zhao, et~al.
\newblock Text2reward: Automated dense reward function generation for reinforcement learning.
\newblock In {\em ICLR}, 2024.

\bibitem[\protect\citeauthoryear{Xu \bgroup \em et al.\egroup }{2024}]{xu2024language}
Zelai Xu, Chao Yu, et~al.
\newblock Language agents with reinforcement learning for strategic play in the werewolf game.
\newblock In {\em ICML}, 2024.

\bibitem[\protect\citeauthoryear{Yao \bgroup \em et al.\egroup }{2024}]{yao2024retroformer}
Weiran Yao, Shelby Heinecke, et~al.
\newblock Retroformer: Retrospective large language agents with policy gradient optimization.
\newblock In {\em ICLR}, 2024.

\bibitem[\protect\citeauthoryear{Zeng \bgroup \em et al.\egroup }{2024}]{zeng2024learning}
Yuwei Zeng, Yao Mu, et~al.
\newblock Learning reward for robot skills using large language models via self-alignment.
\newblock In {\em ICML}, 2024.

\bibitem[\protect\citeauthoryear{Zhai \bgroup \em et al.\egroup }{2024}]{zhai2024fine}
Yuexiang Zhai, Hao Bai, et~al.
\newblock Fine-tuning large vision-language models as decision-making agents via reinforcement learning.
\newblock In {\em NeurIPS}, 2024.

\bibitem[\protect\citeauthoryear{Zhang and Lu}{2024}]{zhang2024adarefiner}
Wanpeng Zhang and Zongqing Lu.
\newblock Adarefiner: Refining decisions of language models with adaptive feedback.
\newblock In {\em NAACL}, 2024.

\bibitem[\protect\citeauthoryear{Zhang \bgroup \em et al.\egroup }{2023a}]{zhang2023large}
Danyang Zhang, Lu~Chen, et~al.
\newblock Large language models are semi-parametric reinforcement learning agents.
\newblock In {\em NeurIPS}, 2023.

\bibitem[\protect\citeauthoryear{Zhang \bgroup \em et al.\egroup }{2023b}]{zhang2023bootstrap}
Jesse Zhang, Jiahui Zhang, et~al.
\newblock Bootstrap your own skills: Learning to solve new tasks with large language model guidance.
\newblock In {\em CoRL}, 2023.

\bibitem[\protect\citeauthoryear{Zhao \bgroup \em et al.\egroup }{2024}]{zhao2024expel}
Andrew Zhao, Daniel Huang, et~al.
\newblock Expel: {LLM} agents are experiential learners.
\newblock In {\em AAAI}, 2024.

\bibitem[\protect\citeauthoryear{Zhou \bgroup \em et al.\egroup }{2024}]{zhou2023large}
Zihao Zhou, Bin Hu, et~al.
\newblock Large language model as a policy teacher for training reinforcement learning agents.
\newblock In {\em IJCAI}, 2024.

\end{thebibliography}

\end{document}